\documentclass[10pt,twocolumn,letterpaper]{article}

\usepackage{cvpr}
\usepackage{times}
\usepackage{epsfig}
\usepackage{graphicx}
\usepackage{amsmath}
\usepackage{amssymb}

\usepackage{makecell}
\usepackage{multirow}
\usepackage{multicol}
\usepackage{lipsum}
\usepackage{xcolor}
\usepackage{booktabs}
\usepackage{tablefootnote}
\usepackage{mathtools}
\usepackage[inline]{enumitem} 
\makeatletter
\newcommand{\inlineitem}[1][]{%
\ifnum\enit@type=\tw@
    {\descriptionlabel{#1}}
  \hspace{\labelsep}%
\else
  \ifnum\enit@type=\z@
       \refstepcounter{\@listctr}\fi
    \quad\@itemlabel\hspace{\labelsep}%
\fi}
\makeatother

\usepackage[breaklinks=true,bookmarks=false,colorlinks]{hyperref}

\cvprfinalcopy 

\def\cvprPaperID{****} 

\ifcvprfinal\fi
\begin{document}

\title{Hit-Detector: Hierarchical Trinity Architecture Search for Object Detection}

\author{
	Jianyuan Guo$^{1,2}$, 
	Kai Han$^{2}$, Yunhe Wang$^{2}$, 
	Chao Zhang$^{1}$\thanks{Corresponding author.}, 
	Zhaohui Yang$^{1}$\\
	Han Wu$^{1}$, Xinghao Chen$^{2}$, Chang Xu$^{3}$
	\\
	\normalsize$^1$ Key Lab of Machine Perception (MOE), Dept. of Machine Intelligence, Peking University.\\
	\normalsize$^2$ Noah's Ark Lab, Huawei Technologies. 
	\normalsize$^3$ School of Computer Science, Faculty of Engineering, University of Sydney.\\
	\small\texttt{\{jyguo, zhaohuiyang, wuhancs\}@pku.edu.cn, chzhang@cis.pku.edu.cn}\\
	\small\texttt{c.xu@sydney.edu.au, \{kai.han, yunhe.wang, xinghao.chen\}@huawei.com} \\ 
}

\maketitle
\thispagestyle{empty}

\begin{abstract}

Neural Architecture Search (NAS) has achieved great success in image classification task. Some recent works have managed to explore the automatic design of efficient backbone or feature fusion layer for object detection. However, these methods focus on searching only one certain component of object detector while leaving others manually designed. We identify the inconsistency between searched component and manually designed ones would withhold the detector of stronger performance. To this end, we propose a hierarchical trinity search framework to simultaneously discover efficient architectures for all components (\ie backbone, neck, and head) of object detector in an end-to-end manner. 
In addition, we empirically reveal that different parts of the detector prefer different operators. Motivated by this, we employ a novel scheme to automatically screen different sub search spaces for different components so as to perform the end-to-end search for each component on the corresponding sub search space efficiently. Without bells and whistles, our searched architecture, namely Hit-Detector, achieves 41.4\% mAP on COCO minival set with 27M parameters. Our implementation is available at \href{https://github.com/ggjy/HitDet.pytorch}{https://github.com/ggjy/HitDet.pytorch}.

\end{abstract}
\section{Introduction}
Object detection is a fundamental task in computer vision and has been widely applied in the real world, such as autonomous vehicles and surveillance video. The advancement of deep learning results in a number of convolutional neural network based solutions of object detection task. Typically, deep learning based detectors can be divided into two categories: (i) one-stage methods including YOLO \cite{yolov1} and SSD~\cite{ssd} which directly utilize CNNs to predict the bounding boxes of interest; and (ii) two-stage approaches such as Faster R-CNN \cite{faster-rcnn} that generates the bounding boxes after extracting region proposals upon a region proposal network (RPN). The advantage of single-stage methods lies in the high detection speed whereas two-stage methods dominate in detection accuracy.

A series of either one-stage~\cite{yolov2,focal,cornernet} or two-stage approaches~\cite{rfcn,mask-rcnn,cascade} have been developed to continuously boost the detection speed and accuracy. However, the manually designed architectures heavily rely on the expert knowledge \cite{detnet} while still might be suboptimal. Thus neural architecture search (NAS) that automates the design of network architectures and minimizes human labor has drawn much attention and made impressive progress, especially in image classification tasks~\cite{nas,pnas,amoebanet,darts,sakai2017semi,mnasnet,fbnet,legonet,ghostnet}. Compared with classification tasks that simply determine \emph{what} the image is, detection tasks need to further figure out \emph{where} the objects are. 
NAS for object detection therefore requires more careful design and is much more challenging. 

\begin{table}[!t]
	\renewcommand\arraystretch{1.0}
	\small
	\centering
	\caption{\small{Comparing our model against some typical two-stage detectors on COCO benchmark. ``4c'' indicates four convolution layers. $^\dagger$ means that NAS-FPN is originally searched for one-stage RetinaNet, we replace the neck in FPN with NAS-FPN to construct the two-stage detector.
	}}
	\label{table:intro comparison}
	\vspace{0.2em}
	\setlength{\tabcolsep}{1mm}
	\resizebox{0.485\textwidth}{!}{
		\hspace{-0.35cm}
		\begin{tabular}{l|c|c|c|c}
			\toprule[1pt]
			\multirow{2}{*}{Model} & {Backbone} & {Neck} & {Head} & mAP \\ 
			& (\#params/M) & (\#params/M) & (\#params/M) & (\%) \\ 
			\hline
			FPN baseline \cite{fpn} & Res50 ({23.5}) & FPN ({3.3}) & 2fc ({14.3}) & 36.2 \\
			NAS-FPN \cite{nasfpn}$^\dagger$ & Res50 ({23.5}) & NAS-FPN ({30.4}) & 2fc ({14.3}) & 38.9 \\
			Backbone baseline & Res50 ({23.5}) & FPN ({3.3}) & 4c1fc ({15.6}) & 36.8 \\
			DetNAS \cite{detnas} & DetNet ({9.4}) & FPN ({2.8}) & 4c1fc ({15.6}) & 40.2 \\ \hline
			NAS-FPN + DetNAS & DetNet ({9.4}) & NAS-FPN ({29.7}) & 4c1fc ({15.6}) & 39.4 \\ \hline
			Hit-Detecotr (ours) & Ours ({13.9}) & Ours ({2.7}) & Ours ({9.9}) & 41.4 \\ 
			\bottomrule[1pt]
	\end{tabular}}
	\vspace{-0.2cm}
\end{table}

Modern object detection systems usually consist of four components: (a) backbone for extracting semantic features, \eg ResNet-50 \cite{resnet} and ResNeXt-101 \cite{resnext}; (b) neck for fusing multi-level features, \eg feature pyramid networks (FPN) \cite{fpn}; (c) RPN for generating proposals (usually in two-stage detector); and (d) head for object classification and bounding box regression. 
Recently, there are works that explore NAS in object detection tasks to search for a good architecture of the backbone~\cite{junran,detnas} or FPN~\cite{nasfpn,autofpn}. With the searched backbone or FPN architectures, these works have achieved higher accuracy than the manually designed baselines with similar numbers of parameters and FLOPs.

Nevertheless, exploiting only one part of the detector at a time cannot fulfill the the potential of each component, and the separately searched backbone and neck may not be optimal or compatible with each other. As shown in Table~\ref{table:intro comparison}, NAS-FPN~\cite{nasfpn} for neck searching achieves a $38.9\%$ mAP which is higher than that of vanilla FPN, and DetNAS~\cite{detnas} for backbone searching outperforms vanilla ResNet-50 backbone with $40.2\%$ mAP. However, a straightforward 
combination of NAS-FPN and DetNAS leads to a worse mAP, \ie, $39.4\%$, let alone outperforms two models. This insightful observation motivates us to take the detector as a whole in NAS.

\begin{figure*}[thb]
	\centering
	\includegraphics[width=\linewidth]{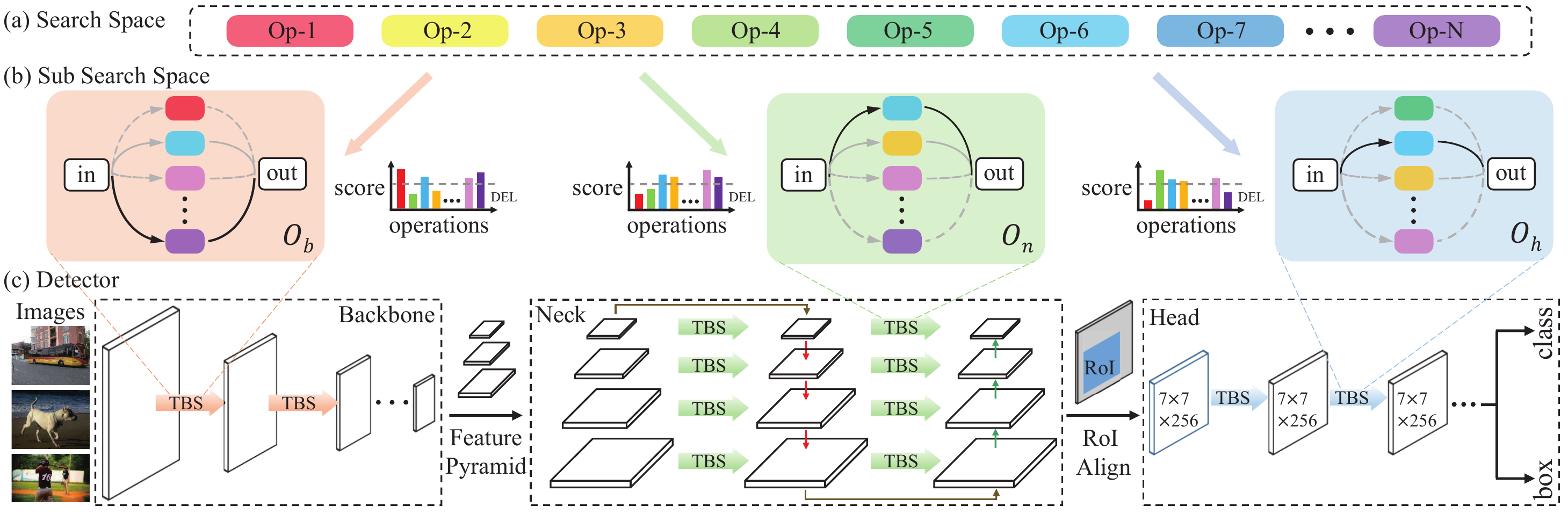}
	\caption{\small{Overview of our Hit-Detector architecture search framework. Our method focuses on searching better architectures of the trinity, \ie backbone, neck, and head for object detector. (a) is the whole search space; (b) indicates three sub search spaces for different components; and (c) shows the end-to-end searching for object detector. ``TBS" denotes the layer to be searched.}}
	\label{fig:hit-detector}
	\vspace{-1em}
\end{figure*}

In this paper, we propose to simultaneously search all components of the detector in an end-to-end manner. Due to the differences among optimum space for each component and the difficulty in optimizing within large search space, we introduce a hierarchical way to mine the proper sub search space from the large volume of operation candidates. In particular, our proposed \emph{Hit-Detector} framework consists of two key procedures as shown in Fig.~\ref{fig:hit-detector}. First, given a large search space containing all the operation candidates, we screen out the customized sub search space suitable for each part of detector with the help of group sparsity regularization. Secondly, we search the architectures for each part within the corresponding sub search space by adopting the differentiable manner. 
Extensive experiments demonstrate that our Hit-Detector achieves state-of-the-art results on the benchmark dataset, which validates the effectiveness of the proposed method.

Our main contributions can be summarized as follows:
\begin{itemize}
	\vspace{-0.1cm}
	\item This is the first time that architectures of backbone, neck and head are searched altogether in an end-to-end manner for object detection.
	\vspace{-0.1cm}
	\item We show that different parts prefer different operations, and propose a hierarchical way to specify appropriate sub search space for different components in detection system to improve the sampling efficiency.
	\vspace{-0.1cm}
	\item Our Hit-Detector outperforms either hand-crafted or automatically searched networks by a large margin with much less computational complexity.
\end{itemize}

\section{Related Work}

Object detection aims at determining \emph{what} and \emph{where} the object is when given an image. Riding the wave of convolutional neural networks, noticeable improvements in accuracy have been made in both one-stage \cite{yolov1,ssd,focal,cornernet,centernet,refinenet} and two-stage \cite{faster-rcnn,mask-rcnn,rfcn,fpn,relation,huang2017speed,cascade,panet} detectors. Generally, object detector consists of four parts: a backbone that extracts features from input images, a neck attached to backbone that fuses multi-level features, a region proposal network which generates prediction candidates on extracted features\footnote{There are only three parts (backbone, neck, and head) for one-stage detector. We take two-stage detector as an example here.}, and a head for classification and localization.

In the past few years, various methods in the literature have been proposed to tackle with detection task and attain significant progress. With NAS prospering automating design of model architecture, it has also boosted the probe into automatically searching for the best architecture for object detection, other than manually design. Here we briefly review some of the recent detectors in two dimensions:

\subsection{Manual design}
The manual design of detectors in mainstream evolution of object detection solutions is promoted by several works. R-CNN \cite{rcnn} is the first to show that a CNN could lead to dramatic performance improvement in object detection. Selective Search \cite{selective_search} is used to generate proposals and SVM is applied to classify each region. Following R-CNN, Fast R-CNN \cite{fast-rcnn} is proposed to improve the speed by sharing computation of convolutional layers between proposals. Faster R-CNN \cite{faster-rcnn} replaces Selective Search with a novel RPN (region proposal network), further promoting the accuracy and makes it possible to train model in an end-to-end manner. In addition, Mask R-CNN \cite{mask-rcnn} extends Faster R-CNN mainly in instance segmentation task. Meanwhile, a series of proposal free detectors, \ie one-stage detectors, has been proposed to speed up the detection. YOLO \cite{yolov1} and YOLOv2 \cite{yolov2} extract features from input images straightly for predicting bound boxes and associated class probabilities through a unified architecture. SSD \cite{ssd} further improves the mAP by predicting a set of bounding boxes from several feature maps with different scales. 

In the mean time, some arts concentrate on improving specific parts such as backbone, neck, and head to improve the efficiency in object detectors. DetNet \cite{detnet} specifically designs a novel backbone network for object detection. FPN \cite{fpn} develops a top-down architecture to effectively encode features from different scales. PANet \cite{panet} further modifies neck module to obtain a better fusion. Focal loss \cite{focal} is proposed to solve the problem of class imbalance. MetaAnchor \cite{metaanchor} proposes a flexible mechanism which generates anchor from arbitrary prior boxes. Light-Head R-CNN \cite{lighthead} designs a light head for two-stage detector to decrease the computation cost accordingly.

\subsection{Neural architecture search}
\paragraph{NAS for image classification}
NAS (neutral architecture search) has attracted great attention recently. Several reinforcement learning based methods \cite{baker2016designing,cai2018efficient,nas,zoph2018learning,pnas} train a RNN controller to generate cell structure and form the network accordingly. Evolutionary algorithm based methods \cite{liu2017hierarchical,ea2019evolving,real2019regularized,real2017large,singlepath,amoebanet,cars} are also proposed to update architectures by mutating current ones. To speed up searching process, gradient based methods \cite{darts,pdarts,fbnet,snas,proxylessnas,xue2019transferable} are proposed for continuous relaxation of search space, which allow the differentiable optimization in architecture search.

\paragraph{NAS for object detection}
In addition to NAS works on classification, some recent works attempt to develop NAS for object detector. NATS~\cite{junran} claims that the effective receptive field of backbone is critical and uses NAS to search different dilation rates for each convolution layer in backbone. Similarly, DetNAS~\cite{detnas} aims to search a better backbone for detection task. NAS-FPN~\cite{nasfpn} targets at a better architecture of feature pyramid network for object detection, adopting NAS to discover a new feature pyramid architecture covering all cross-scale connections. Auto-FPN~\cite{autofpn} sequentially searches a better fusion of the multi-level features for neck and a better structure for head. However, there are two shortcomings in above methods: (i) the search space is defined by human prior and might be too naive for searching (\eg four choices based on ShuffleNetV2 \cite{shufflenetv2} in DetNAS \cite{detnas}); (ii) in each work, only one certain part is searched (\eg backbone in \cite{junran,detnas}, neck in \cite{nasfpn}) can lead to suboptimal result in detection task. To tackle these two challenges, we propose Hit-Detector that filters proper search space for each parts hierarchically and searches every parts for a better detector in an end-to-end manner.


\section{Hit-Detector}
In this section, we introduce the proposed Hierarchical Trinity architecture search algorithm for object detection and the resulted detector, \ie Hit-Detector. We first identify and analyze the problems in current NAS algorithms for object detection to clarify our motivation that we need to search all components together. Then we detail how to hierarchically filter the sub search space for each parts, and last, the end-to-end search process for Hit-Detector is depicted. The following statement of search algorithm is based on two-stage detection methods and can be easily applied to one-stage methods.

\subsection{Preliminaries and Motivation}
Two-stage detection system can be decoupled as four components: (i) \textbf{Backbone.} Commonly used backbones in detection system such as ResNet \cite{resnet} and ResNeXt \cite{resnext} are mostly manually designed for classification tasks. Usually, the largest proportion of parameters in a detector comes from backbone. For example, ResNet-101, the backbone of FPN~\cite{fpn}, takes up 71\% parameters of all, leaving a large potential for searching; (ii) \textbf{Neck.} Employing in-network feature pyramids to approximate different receptive fields can help detector localize objects better. Previous architectures of feature fusion neck are manually designed while NAS can exploit better operations for merging features from different scales; (iii) \textbf{RPN.} The typical RPN is lightweight and efficient: a convolution layer followed by two fully connected layers for region proposal classification and bounding box regression. We follow this design for its efficiency and do not search for this part; and (iv) \textbf{Head.} Detectors usually have a heavy head attached to former network. For example, Faster R-CNN \cite{faster-rcnn} employs the 5-th stage in ResNet \cite{resnet} and FPN \cite{fpn} uses two large fully connected layers (34\% parameters of whole detector) to execute classification and regression, which are inefficient for detection. We call the backbone, the neck, and the head which are valuable to be searched as detector trinity in this paper.

Recently, several methods have been proposed to search either backbone~\cite{detnas} or feature pyramid architecture~\cite{nasfpn} for object detection. Denoting the search space $\mathcal{A}$ as a directed acyclic graph (DAG) where the nodes indicate the features while directed edges are associated with multifarious operations such as convolution layer and pooling layer. We can view each path from start point to end point in the graph as an architecture $\alpha\in\mathcal{A}$. Previous NAS works for object detection can be formulated as the optimization problem:
\begin{equation}\label{eq:onenas}
\begin{aligned}
\alpha^*
&=\mathop{\arg\min}\limits_{\alpha\in\mathcal{A}}f(\alpha)=
\mathop{\arg\min}\limits_{\alpha\in\mathcal{A}}\mathcal{L}^{det}_{val}(\alpha,w^*(\alpha))\\
&=\mathop{\arg\min}\limits_{\alpha\in\mathcal{A}}\mathcal{L}^{det}_{val}(\alpha,\mathop{\arg\min }\limits_{w}\mathcal{L}^{det}_{train}(\alpha,w)).
\end{aligned}
\end{equation}
The purpose of NAS process is to find a specific architecture $\alpha\in\mathcal{A}$ that minimizes the validation loss $\mathcal{L}^{det}_{val}(\alpha,w^*_\alpha)$ with the trained weights $w^*_\alpha$. The above formulation can represent searching on backbone (\eg $\mathcal{A}$ denotes the backbone search space) or feature pyramid network (\eg $\mathcal{A}$ denotes the FPN search space).

However, the backbone, the neck (feature fusion network), and the head in object detection system should be highly consistent to each other. Only redesigning the backbone or neck is not enough, which can lead to suboptimal results. As shown in Table~\ref{table:intro comparison}, combining the separately searched backbone in DetNAS and neck in NAS-FPN leads to a worse result. We claim that separately search backbone $\alpha$, neck $\beta$, and head $\gamma$ in detector is inferior to search all these components end-to-end:
\begin{equation}\small
\begin{aligned}
& \mathcal{L}^{det}_{val}(\alpha',\beta',\gamma') \geq \mathcal{L}^{det}_{val}(\alpha^*,\beta^*,\gamma^*),\\
\!\mathrm{s.t.} ~& \alpha',\beta',\gamma'\!=\!\mathop{\arg\min}\limits_{\alpha}f(\alpha),\mathop{\arg\min}\limits_{\beta}f(\beta),\mathop{\arg\min}\limits_{\gamma}f(\gamma),\\
&\alpha^*,\beta^*,\gamma^*=\mathop{\arg\min}\limits_{\alpha,\beta,\gamma}\mathcal{L}^{det}_{val}(\alpha,\beta,\gamma,w^*(\alpha,\beta,\gamma)),\\
&\alpha\in\mathcal{A}_b,~ \beta\in\mathcal{A}_n,~ \gamma\in\mathcal{A}_h.
\end{aligned}
\end{equation}
where $\alpha',\beta',\gamma'$ are obtained by solving corresponding optimization problem as in Eq.~\ref{eq:onenas} separately, while $\alpha^*,\beta^*,\gamma^*$ are optimized through the end-to-end searching algorithm, and $\mathcal{A}_b,\mathcal{A}_n,\mathcal{A}_h$ are search spaces for backbone, neck, and head, respectively.

In this paper, we propose to search the trinity in detectors, \ie backbone, neck, and head in an end-to-end manner:
\begin{equation}\label{eq:end-to-end}
\begin{aligned}
&\alpha^*,\beta^*,\gamma^*
=\mathop{\arg\min}\limits_{\alpha,\beta,\gamma}\mathcal{L}^{det}_{val}(\alpha,\beta,\gamma,w^*(\alpha,\beta,\gamma))=\\
&\mathop{\arg\min}\limits_{\alpha,\beta,\gamma}\mathcal{L}^{det}_{val}(\alpha,\beta,\gamma,\mathop{\arg\min }\limits_{w}\mathcal{L}^{det}_{train}(\alpha,\beta,\gamma,w)).
\end{aligned}
\end{equation}
As shown in Figure~\ref{fig:hit-detector}, we propose the hierarchical trinity architecture search framework to solve problems in Eq.~\ref{eq:end-to-end}, which includes two procedures: screening sub search space for each component and searching the trinity end-to-end.

\subsection{Screening Sub Search Space}
\begin{figure}[t]
	\centering
	\includegraphics[width=\linewidth]{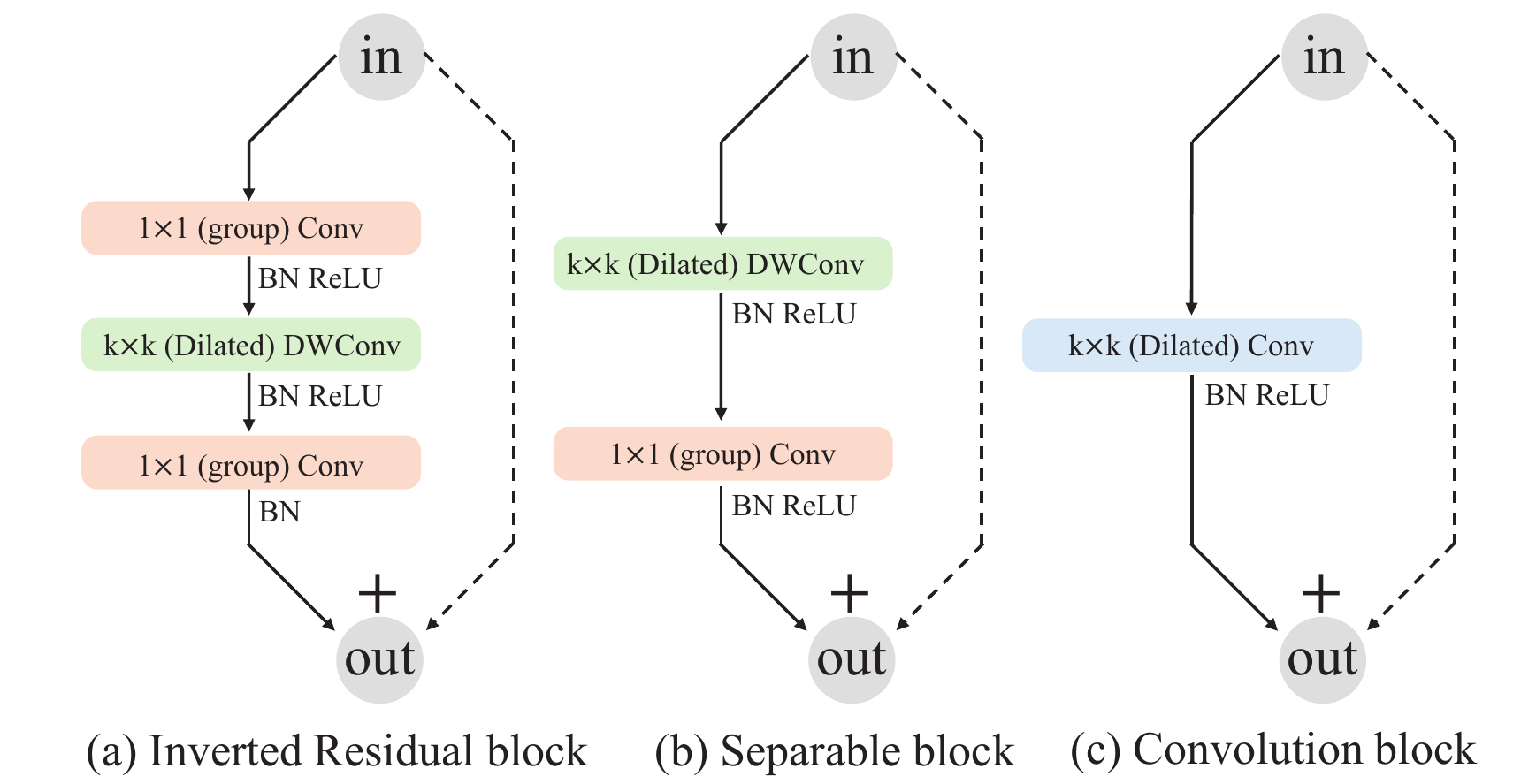}
	\caption{\small{The block structure of our search space, \eg, candidate operations in inverted residual block can choose a different expansion rate, kernel size and number of groups for group convolution.}}
	\label{fig:search space}
\end{figure}

Search space is one of the key factors in neural architecture search. The search spaces in \cite{detnas,junran,autofpn} are manually designed with a limited number of operation candidates. Furthermore, there is usually the case that manually designed search space is not suitable for the architecture that needs to be optimized. In order to prevent insufficient search space, we start from the FBNet \cite{fbnet} and traverse as many candidates as possible to form a large set of operation candidates, which are illustrated in Figure~\ref{fig:search space}. Inverted residual block \cite{mobilenetv2} contains a $1\times1$ convolution, a $k\times k$ depthwise convolution, another $1\times1$ convolution and an expansion factor $e$. Separable block contains a $k\times k$ depthwise convolution and a $1\times1$ convolution. If the output dimension is the same as input's, we use a skip connection to add them together.

As shown in Figure~\ref{fig:hit-detector}(a), the whole search space consists of $N$ different operation candidates, \eg, $N=32$ in our experimental setting. If we directly apply this large search space for NAS, the memory and computational overhead is so heavy that ordinary hardware cannot support the search process efficiently. Moreover, in Sec~\ref{sec:ablation}, we empirically show that the same operation can have different impacts on final result at different parts in detection system. In order to find the most suitable search space for each component and reduce the computational burden, we propose a screening scheme to hierarchically filter operation candidates for each component. As shown in Figure~\ref{fig:hit-detector}(b), every candidate is associated with a score. The candidates with higher scores are retained, and others with lower scores are deleted.

Take the search space of backbone as an example, we conduct layer-wise search where each layer can choose any operation from the candidates. Assuming that the backbone has $L$ layers, the score of the $i$-th operation in the $l$-th layer is denoted as $\alpha_{l,i}$. All scores form the architecture parameter matrix $\alpha\in\mathbb{R}^{L\times N}$, where the $i$-th column represents the scores of the $i$-th operation in corresponding layers. With a relatively large amount of candidates, the scores tend to be similar and are hard to distinguish. In order to screen the most suitable operation subset, the matrix $\alpha$ is imposed with the column-sparse regularization:

\begin{equation}
\label{eq:regularization}
\mathop{\min}\limits_{\alpha} f(\alpha)+\mu\mathop{\min}\limits_{i}(\sqrt{\sum\nolimits_{l=1}^{L}\alpha_{l,i}^2}),
\end{equation}
$\mu$ is a trade-off hyper-parameter. In screening stage, the architecture parameter $\alpha$ is learned and the scores of candidates are ranked accordingly, the last several candidates will be removed from search space gradually until a search space with size $N_{b}$ is obtained. The sub search spaces for backbone, neck, and head are hierarchically selected from the whole search space, namely, $\mathcal{O}_{b}$, $\mathcal{O}_{n}$, and $\mathcal{O}_{h}$.

\subsection{End-to-end Search in Hit-Detector}
After obtaining the suitable sub search space for each component, we start the end-to-end search for object detector. We adopt the differentiable manner proposed in \cite{darts} to solve Eq.~\ref{eq:end-to-end}, and represent the sub search space by a stochastic supernet. During searching, each intermediate node is computed as a weighted sum based on all candidates. For backbone, the $l$-th node is formulated as
\begin{align}\label{eq:backbone-node}
x_{l}=\sum_{o\in\mathcal{O}_{b}} {\frac{\exp(\alpha_{l}^o)}{\sum_{o'\in\mathcal{O}_{b}}{\exp(\alpha_{l}^{o'})}} o(x_{l-1})},
\end{align}
where $x_{l}$ is the output of the $l$-th layer, $\alpha_{l}^o$ is the parameter for operation $o(\cdot)$, and $\mathcal{O}_{b}$ indicates the sub search space for backbone. The nodes in neck and head are similar to Eq.~\ref{eq:backbone-node}. This continuous relaxation makes the entire framework differentiable to both operation weights and architecture parameters, so that we can perform architecture search in an end-to-end manner. In testing phase, we can easily decode architectures from $\alpha,\beta,\gamma$ by choosing operation with the highest score in each layer and construct detector using the selected operation. 

The details of our detector supernet are described as follows. The basic structure of \textbf{backbone} consists of a 3$\times$3 convolution head with stride of 2 followed by four stages that contain $4+4+8+4=20$ blocks to be searched. Conventionally, we define the layers that producing feature map with the same spatial size belong to the same network stage. In each stage, the first block has a stride of 2 for downsampling, and the last feature map generated by the backbone has a downsample rate of 32 compared to the input image. The channel of each stage is set to $\{48,96,256,352\}$ respectively. We use $\{C_1,C_2,C_3,C_4\}$ to represent feature maps of stride $\{4,8,16,32\}$ generated by backbone. Then we send the feature pyramid into the \textbf{neck}. Generally, high level features have better semantic information while low level features have more accurate location information. To propagate semantic signals and accurate information about localization to features from lower and higher level, we use both top-down and bottom-up path augmentation to enhance features from different levels inspired by \cite{panet}. We use $\{P_1,P_2,P_3,P_4\}$ and $\{N_1,N_2,N_3,N_4\}$ to denote feature maps after top-down and bottom-up path, respectively. The whole neck contains $4+4=8$ lateral connections to be searched, thus each feature level can search for different operations to assign proper receptive fields. Given the aligned feature maps generated by region proposal network, \textbf{head} is applied to predict final classification and refine the bounding box of the object. We design $4$ blocks to be searched followed by a fully connected layer to form the detection head. The number of the output channels for each level in neck and head is 256 and the output of the fully connected layer is a 512 dimensional vector. In our experiments, we set ${N_b}={N_n}={N_h}=8$. Even if we extract the sub search spaces carefully, the final search space contains $8^{(20+8+4)}\approx 7.9\times10^{28}$ possible architectures.

\subsection{Optimization}
\label{sec:optimization}
In order to control the computational cost of the searched detector at the same time, we add the FLOPs constraint as a regularization term in the loss function and rewrite the Eq.~\ref{eq:end-to-end} as:
{\small
\begin{align}
\label{eq:resource constraint}
\mathop{\min}\limits_{\alpha,\beta,\gamma}\mathcal{L}^{det}_{val}(\alpha,\beta,\gamma,w^*(\alpha,\beta,\gamma)) \!+\! \lambda({\rm C}(\alpha)\!+\!{\rm C}(\beta)\!+\!{\rm C}(\gamma)))
\end{align}}
where $\lambda$ is the coefficient to balance accuracy and cost of the detector, ${\rm C}(\alpha)$ indicates FLOPs of the backbone part and can be decomposed as linear sum of each operations:
\begin{align}
{\rm C}(\alpha) 
=\sum_{l}\sum_{\mathclap{o\in \mathcal{O}_{b}}}{\alpha_l^{o}{\rm {FLOPs}}(o,l)},
\end{align}
${\rm C}(\beta)$ and ${\rm C}(\gamma)$ can be calculated similarly.

It is clear that continuous relaxation from Eq.~\ref{eq:backbone-node} is differentiable with respect to architecture parameters and operation weights, thus $\{\alpha, \beta, \gamma, w\}$ can be optimized jointly using stochastic gradient descent. We adopt first-order approximation following \cite{darts} and update architecture parameters and operation weights alternately. We first fix $\{\alpha, \beta, \gamma\}$ and compute $\partial\mathcal{L}/\partial w$ to train the network weights on 50\% training data, then we fix the network weights and calculate $\partial\mathcal{L}/\partial\alpha$, $\partial\mathcal{L}/\partial\beta$, and $\partial\mathcal{L}/\partial\gamma$ to update the architecture parameters on the remaining 50\% training data, where the loss function $\mathcal{L}$ is the localization and classification losses calculated on the detection mini-batch. The optimization alternates until the supernet converges.

\section{Experiments}

In this section, we investigate the effectiveness of proposed Hit-Detector by conducting elaborate experiments on COCO benchmark.

\begin{figure*}[t]\centering
	\centering
	\includegraphics[width=\linewidth]{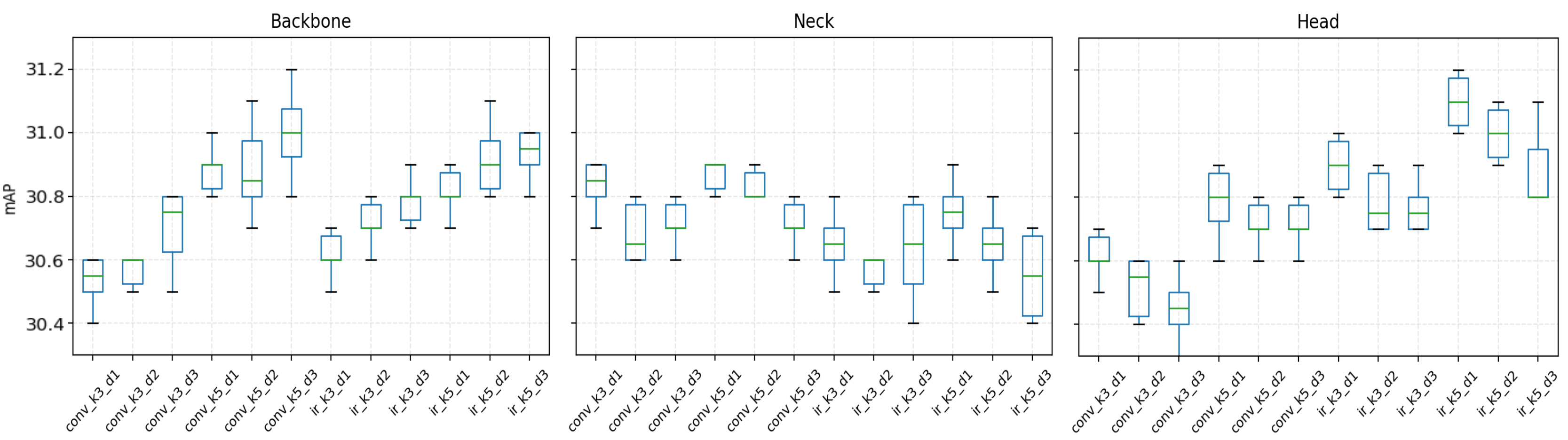}
	\caption{\small{Influence of replacing a certain operation in different parts of the detector on COCO minival. FPN (4clfc head) is taken as a base detector and the input image is of size 320$\times$320. 
	Taking the left figure as an example, each time one layer from backbone is randomly chosen to be replaced by an operation candidate. For one operation candidate, such random process is repeated for 6 times. 
	}}
	\label{fig:toy example}
	\vspace{-1em}
\end{figure*}

\subsection{Dataset and Metrics}
We conduct experiments on MS COCO 2014 dataset \cite{coco}, which contains 80 object classes. Following \cite{bell2016inside,fpn}, our training set is the union of 80k training images and 35k subset of validation images (trainval35k), and validation set is the remaining 5k validation images (minival). We consider Average Precision with different IoU thresholds from 0.5 to 0.95 with an interval of 0.05 as evaluation metric, \ie, $\rm mAP$, $\rm {AP}_{50}$, $\rm {AP}_{75}$, $\rm {AP}_{S}$, $\rm {AP}_{M}$ and $\rm {AP}_{L}$. The last three measure performance \wrt objects with different scales.

\subsection{Implementation Details}
Our implementation is based on mmdetection \cite{mmdetection} with Pytorch framework \cite{pytorch}. We firstly filter three sub search spaces for backbone, neck, and head, respectively. Then we search for our Hit-Detector following the algorithm in sec~\ref{sec:optimization}. Finally we train our searched model on the training set mentioned above. We use only horizontal flipping as data augmentation for training, and there is no data augmentation for testing. The experiments are conducted on 8 V100 GPUs.

\vspace{-0.1cm}
\paragraph{Screening sub search space.}
We sequentially screen sub search spaces for different parts of the detector due to the GPU memory constraint. The number of operations in the whole search space $\mathcal{O}$ is 32 as shown in Fig.~\ref{fig:search space}, and the number of operations to be screened for all three sub search spaces $\mathcal{O}_b$, $\mathcal{O}_n$, and $\mathcal{O}_h$ is set as 8. We halved the depth of backbone supernet to $2+2+4+2=10$ for simplifying this process. We first pretrain the backbone supernet on ImageNet for 10 epochs with fixed architecture parameters, then fine-tune the entire detector supernet on COCO. SGD optimizer with momentum 0.9 and cosine schedule with initial learning rate 0.04 is used for learning model weights, while Adam optimizer with learning rate 0.0004 is adopted for updating architecture parameters. The $\mu$ in Eq.~\ref{eq:regularization} is empirically set to 0.1. We begin to optimize architecture parameters at the 6-th epochs and finish searching at 12 epochs, and we don't use resource constraint in this stage.

\vspace{-0.1cm}
\paragraph{Trinity Architecture searching.}
After screening sub search spaces, we start to search the backbone, neck and head in an end-to-end manner. We pretrain the new backbone supernet on ImageNet based on the corresponding sub search space, and then search the detector on COCO dataset. The optimizers to learn the architecture parameters and weights are the same as we do when screening sub search spaces. The $\lambda$ in Eq.~\ref{eq:resource constraint} is empirically set to 0.01 to trade off the accuracy and the FLOPs constraint.

\vspace{-0.1cm}
\paragraph{Training details.}
We first pretrain the searched backbone on ImageNet for 300 epochs, then we fine-tune the whole detector on COCO training set. The input image is resized such that its shorter side has 800 pixels. We use SGD optimizer with a batch size of 4 images per GPU, and our model is trained for 12 epochs, known as 1$\times$ schedule. The initial learning rate is 0.04 and is divided by 10 at the 8-th and 11-th epoch. We set momentum as 0.9 and weight decay as 0.0001.

\begin{table*}[h]\small
	\renewcommand\arraystretch{1.0}
	\begin{center}
	\caption{\small{Comparisons of the number of parameters, FLOPs and mAP on COCO minival. The FLOPs is based on the 800$\times$1200 input and 1000 proposals in region proposal network. $^\ddagger$ means the 2x schedule in training.
	}}
	\label{table:main result}
	\vspace{0.1em}
	\begin{tabular}{l|ccc|cc|c|ccccc}
		\toprule[1pt]
		\multirow{2}{*}{model} & \multicolumn{3}{c|}{modified} & \# params & \# FLOPs & \multirow{2}{*}{mAP} & \multirow{2}{*}{$\rm {AP}_{50}$} & \multirow{2}{*}{$\rm {AP}_{75}$} & \multirow{2}{*}{$\rm {AP}_{S}$} & \multirow{2}{*}{$\rm {AP}_{M}$} & \multirow{2}{*}{$\rm {AP}_{L}$}\\
		& B & N & H & (total) & (total) & & & & \\ \hline
		FPN \cite{fpn} & - & - & - & 41.76M & 197.4B & 36.2 & 58.0 & 39.1 & 21.3 & 40.0 & 46.1 \\
		MobileNetV2 \cite{mobilenetv2} & $\checkmark$ &  &  & 19.61M & 116.94B & 30.1 & 51.8 & 30.9 & 16.7 & 33.0 & 38.7 \\
		ResNeXt-101 \cite{resnext} & $\checkmark$ &  &  & 60.38M & 273.3B & 40.3 & 62.1 & 44.1 & 23.6 & 45.0 & 51.6 \\ 
		\hline
		FBNet-C \cite{fbnet} & $\checkmark$ &  &  & 21.40M & 119.0B & 35.1 & 57.4 & 37.2 & 19.3 & 38.3 & 46.7 \\ 
		DetNAS-1.3G \cite{detnas} & $\checkmark$ &  & $\checkmark$ & 28.45M & 254.1B & 40.2 & 61.5 & 43.6 & 23.3 & 42.5 & 53.8 \\
		NASFPN \cite{nasfpn} &  & $\checkmark$ &  & 68.86M & 616.9B & 38.9 & 59.3 & 42.3 & 22.3 & 42.8 & 49.8 \\
		DetNAS-1.3G + NASFPN & $\checkmark$ & $\checkmark$ & $\checkmark$ & 55.29M & 672.9B & 39.4 & 59.6 & 42.1 & 23.7 & 43.2 & 50.4 \\ 
		NATS-C \cite{junran} &  $\checkmark$ &  &  & 41.76M & 197.4B & 38.4 &  61.0 & 41.2 & 22.5 & 41.8 & 50.4 \\
		Auto-FPN$^\ddagger$ \cite{autofpn} & & $\checkmark$ & $\checkmark$ & 32.64M & 476.6B & 40.5 & 61.5 & 43.8 & \textbf{25.6} & 44.9 & 51.0 \\ \hline
		Hit-Detector & $\checkmark$ &  $\checkmark$ & $\checkmark$ & 27.12M & 272.3B & \textbf{41.4} & \textbf{62.4} &  \textbf{45.9} & 25.2 & \textbf{45.0} &  \textbf{54.1} \\
		\bottomrule[1pt]
	\end{tabular}
	\end{center}
\end{table*}

\subsection{Main Results}
\paragraph{Comparisons with hand-crafted methods.}
FPN \cite{fpn} with ResNet-50 as backbone is the baseline model here. We replace the backbone in FPN with other excellent backbones, \ie MobileNetV2 \cite{mobilenetv2} and ResNeXt-101 \cite{resnext}, and form two competitor models accordingly. As shown in Table~\ref{table:main result}, Hit-Detector surpasses baseline by a large margin with much less parameters. In addition, Our method is 1.1\% higher on mAP compared to the ResNeXt based detector with less than one half of the parameters. And we outperform MobileNetV2 by 11.3\% on mAP with only a bit more parameters. This demonstrates that our method can find a better architecture than hand-crafted baselines.

\vspace{-0.1cm}
\paragraph{Comparisons with NAS based methods.}
As shown in Table~\ref{table:main result}, we compare our method with both detector searched on COCO benchmark and detector that adopts NAS based model as backbone. FBNet \cite{fbnet} is searched on ImageNet dataset and we directly apply it as the backbone of a detector, however, its performance on detection task is disappointing. DetNAS \cite{detnas} aims to search a better backbone directly on detection benchmark while leaves the rest parts unchanged. Our method outperforms DetNAS by 1.2\% with less parameters and pretty much FLOPs. It can be seen that Hit-Detector surpasses all previous NAS based methods, which indicates that it is important to search the trinity in an object detector.

\vspace{-0.1cm}
\paragraph{Comparison with State-of-the-arts on test-dev set.}
We also compare the results of our Hit-Detector with other state-of-the-art methods on the COCO test-dev, and we summarize the comparisons in Table~\ref{table:test-dev}. Hit-Detector only applies horizontal flipping as data augmentation and 2x training scheme, achieves 44.5\% mAP without bells and whistles. Our model has less parameters and performs even better compared to other detectors such as TridentNet and NAS-FPN. This demonstrates that our method can find a better architecture than hand-crafted or partly searched methods. 

\begin{table}[thb]\small
	\renewcommand\arraystretch{1.}
	\centering
	\caption{\small{Comparisons of single-model results on COCO test-dev.}}
	\label{table:test-dev}
	\vspace{0.1em}
	\begin{tabular}{l|c|c}
		\toprule[1pt]
		model & test size & mAP \\ \hline
		R-FCN \cite{rfcn} & 600/1000 & 32.1 \\
		Faster R-CNN \cite{faster-rcnn} & 600/1000 & 30.3 \\
		Deformable \cite{deformable} & 600/1000 & 34.5 \\
		FPN \cite{fpn} & 800/1200 & 36.2 \\
		Mask R-CNN \cite{mask-rcnn} & 800/1200 & 38.2 \\
		RetinaNet \cite{focal} & 800/1200 & 39.1 \\
		Light head R-CNN \cite{lighthead} & 800/1200 & 41.5 \\ 
		PANet \cite{panet} & 800/1000 & 42.5 \\
		TridentNet \cite{trident} & 800/1200 & 42.7 \\
		NAS-FPN \cite{nasfpn} & 1024/1024 & 44.2 \\
		\hline
		Hit-Detector & 800/1200 & 44.5 \\
		\bottomrule[1pt]
	\end{tabular}
\end{table}

\subsection{Ablation Studies}
\label{sec:ablation}
\paragraph{Influence of different operations.} We use a toy example to demonstrate that different parts of the detector are sensitive to different operations, so that different parts need different sub search spaces. Here we simply choose 12 operations from the whole search space as shown in Figure~\ref{fig:search space}. For example, ``conv\_k3\_d1" denotes the convolution block with kernel size 3 and dilation rate 1, and ``ir\_k3\_d1" denotes the inverted residual block with kernel size 3 and dilation rate 1, respectively. Take the left figure as an example, for each operation candidate, we randomly select a layer from backbone and replace original operation with the operation candidate to explore the influence of the selected operation candidate. For each part (backbone, neck, and head), we repeat the random process 6 times to ensure that operation candidate can be inserted in layers of different depths. We can find that (i) different parts prefer different operations. For example, operation ``conv\_k5\_d3" achieves the highest mAP in backbone while ``conv\_k5\_d1" performs the best in head; (ii) one operation has different performances in different parts, ``conv\_k3\_d1" attains 30.4\%-30.6\% mAP in backbone but gets 30.7\%-30.9\% mAP in neck; and (iii) the performance of one operation within the same part is stable enough so that it's reasonable for us to screen sub search spaces by different parts.

\begin{figure}[t]
	\centering
	\includegraphics[width=.49\linewidth]{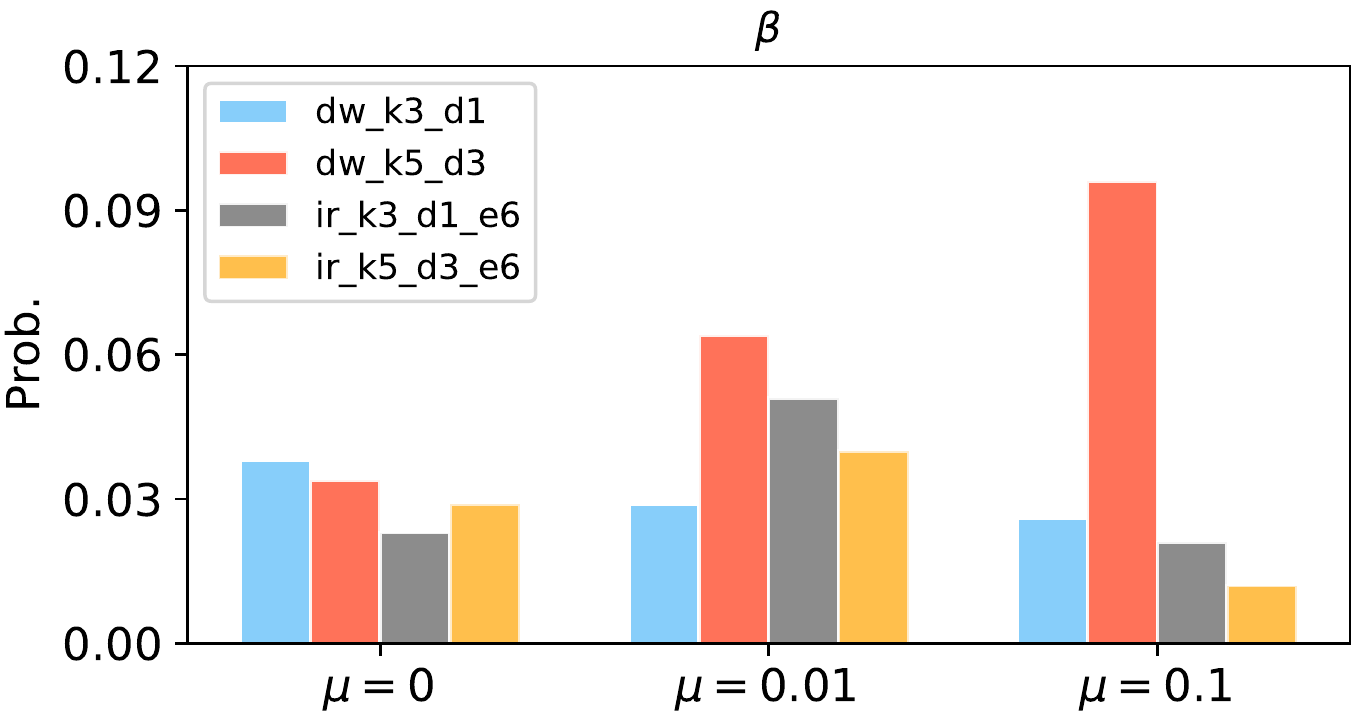}
	\includegraphics[width=.49\linewidth]{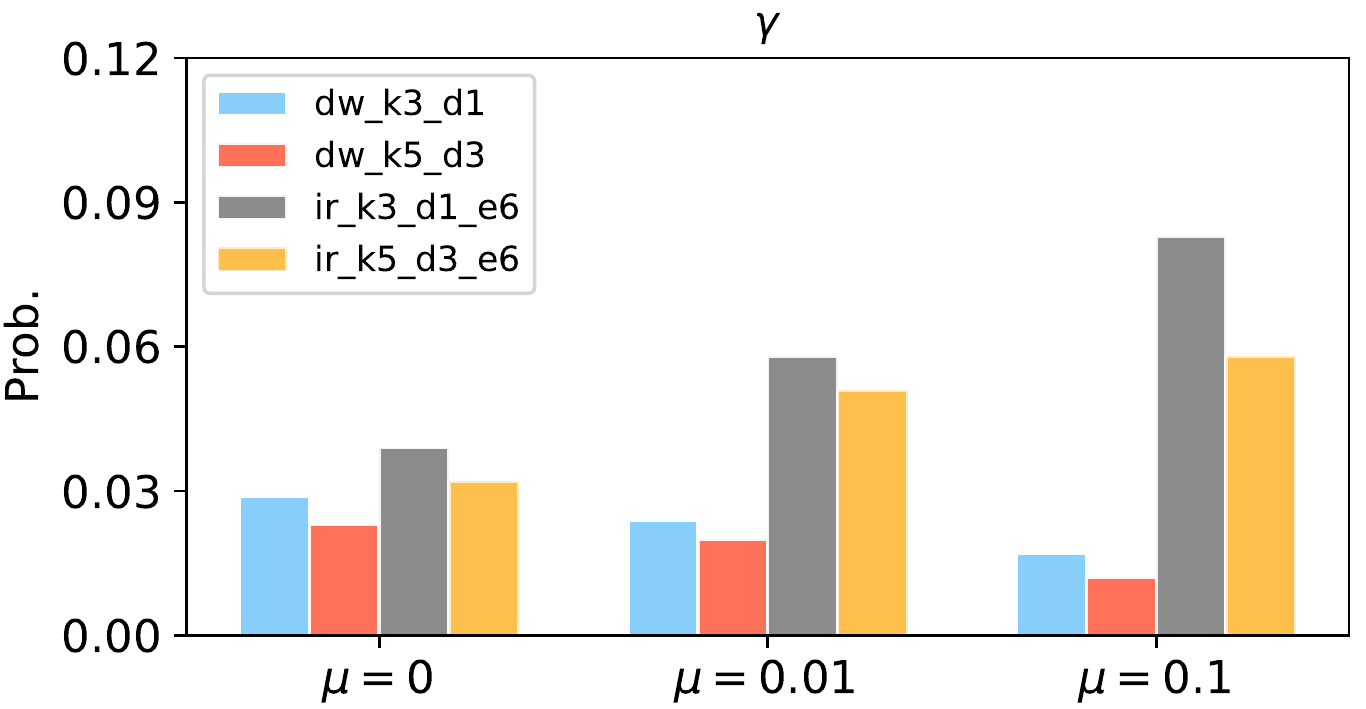}
	\caption{\small{Ablation study of the trade-off parameter $\mu$ in column-sparse regularization.}}
	\label{fig:regularization}
	\vspace{-0.cm}
\end{figure}

\vspace{-0.1cm}
\paragraph{Column-sparse regularization.}
To further evaluate the influence of the column-sparse regularization, we set the $\mu$ in Eq.~\ref{eq:regularization} to $\{0,0.01,0.1\}$ and randomly choose 4 operation candidates to draw the Figure~\ref{fig:regularization}. We can find that if there is no column-sparse regularization ($\mu=0$), probabilities of different operations are rather similar, which makes the screening process unstable. As $\mu$ increases, differences between operations become more significant, which helps to screen proper sub search space easier. We set $\mu=0.1$ in the rest of our experiments.

\vspace{-0.1cm}
\paragraph{Importance of screening sub search space.}
We study the influence of screening different sub search spaces through the ablation study shown in Table~\ref{table:subspace}. When three parts have the same sub search space, the mAP decreases from 41.4\% to 40.1\%, which indicates that different parts need to have their proper sub search space to achieve better performance.

\begin{figure}[thb]
	\centering
	\includegraphics[width=\linewidth]{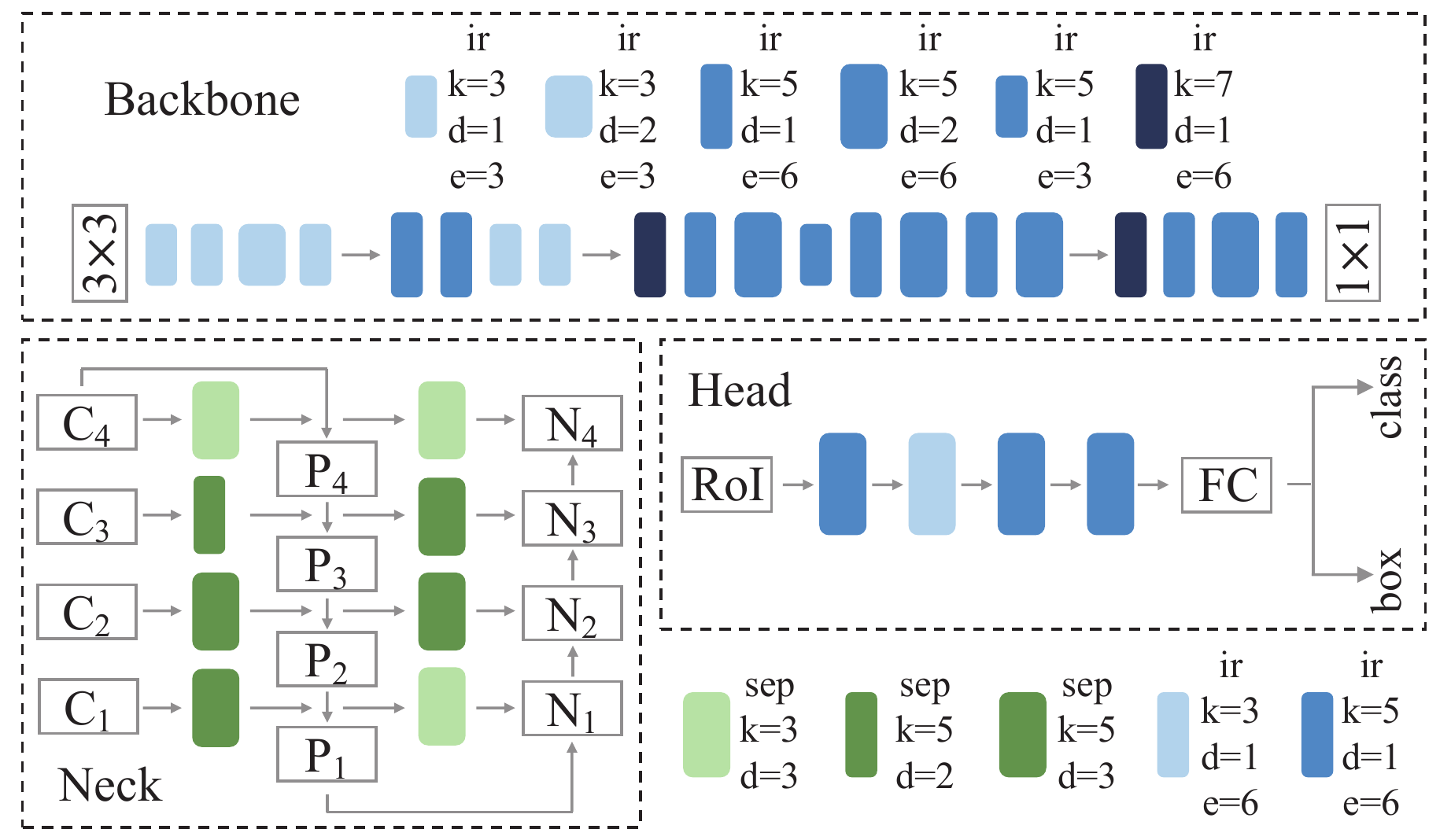}
	\caption{\small{Visualization of the searched Hit-Detector. We use rectangle boxes to denote operations for each layer.}}
	\label{fig:searched_arch}
	\vspace{-0.1cm}
\end{figure}

\begin{table}[h]\small
	\renewcommand\arraystretch{1.05}
	\centering
	\caption{\small{Ablation study of screening sub search space.}}
	\label{table:subspace}
	\vspace{0.1em}
	\begin{tabular}{l|ccc}
		\toprule[1pt]	
		model & mAP & $\rm {AP}_{50}$ & $\rm {AP}_{75}$ \\ \hline
		1 sub search space & 40.1 & 61.2 & 44.0 \\ 
		3 sub search space & 41.4 & 62.4 & 45.9 \\ 
		\bottomrule[1pt]
	\end{tabular}
\end{table}

\paragraph{Importance of trinity architecture searching.}
We explore the impact of each searched part in Hit-Detector here. Taking ResNet-50 based FPN \cite{fpn} as baseline model, we replace one part of baseline model with our searched component each time and verify the new model on COCO minival. As shown in Table~\ref{table:albation trinity}, our searched backbone architecture achieves 39.2\% mAP, and the searched head also attains 38.5\% mAP. With the searched trinity, the mAP of Hit-Detector increases to 41.4\%, which is much higher than baseline model and the single part competitor. Another noteworthy finding is that the backbone and head can boost performance better than neck. We think the main reasons are that (i) object detection emphasizes more on perception to the location of each object in an image, thus the backbone designed for detection can perform better compared to the one designed for classification; and (ii) head aims to identify and refine the location of bounding boxes, thus searching more suitable convolution layers in head can bring more benefits to detection task.

Our searched Hit-Detector is depicted in Figure~\ref{fig:searched_arch}. We observed that the backbone prefers operations that have large kernel size. However, convolution layer with dilation 3 which achieves the best mAP in Figure~\ref{fig:toy example} is not chosen. One main reason is that the ResNet-50 backbone in toy example only includes $3\times3$ convolution, thus operation with dilation 3 can boost the performance prominently. Except that, backbone and head prefer operations with bigger expansion rate such as inverted residual block which has more intermediate channels to increase the expression of feature, while the neck prefers bigger dilation rate for larger receptive fields.

\begin{table}[h]\small
	\renewcommand\arraystretch{1.05}
	\centering
	\caption{\small{Evaluations of different parts searched in Hit-Detector. $\checkmark$ means we replace the part in baseline with the searched one.}}
	\label{table:albation trinity}
	\vspace{0.1em}
	\begin{tabular}{l|ccc|c}
		\toprule[1pt]
		\multirow{2}{*}{model} & \multicolumn{3}{c|}{Searched} & \multirow{2}{*}{mAP} \\  
		& backbone & neck & head & \\ \hline
		FPN baseline &  &  &  & 36.2 \\ 
		4c1fc baseline &  &  & $\checkmark$ & 36.8 \\ 
		\hline
		Searched backbone & $\checkmark$ &  &  & 39.2 \\ 
		Searched neck &  & $\checkmark$ &  & 37.4 \\ 
		Searched head &  &  & $\checkmark$ & 38.5 \\ \hline
		Hit-Detector & $\checkmark$ & $\checkmark$ & $\checkmark$ & 41.4 \\ 
		\bottomrule[1pt]
	\end{tabular}
\end{table}
 
\paragraph{Extension to one-stage detector.}
To evaluate the generalization of our method, we apply it to RetinaNet \cite{focal} for searching one-stage detector. The search algorithm is the same as mentioned in sec~\ref{sec:optimization}. As shown in Table~\ref{table:one-stage}, our model outperforms RetinaNet by 1.3\% in terms of mAP and the model size is smaller than both VGG based SSD and ResNet-50 based RetinaNet.

\begin{table}[ht]\small
	\renewcommand\arraystretch{1.05}
	\centering
	\caption{\small{Extending Hit-Detector to one-stage detector on COCO minival (1x schedule). $^\dagger$ denotes the result of our implementation.}}
	\vspace{0.1em}
	\label{table:one-stage}
	\setlength{\tabcolsep}{1.5mm}
	\begin{tabular}{l|c|ccc}
		\toprule[1pt]
		model & \# Params & mAP & $\rm {AP}_{50}$ & $\rm {AP}_{75}$ \\ \hline
		SSD-VGG19 \cite{ssd} & 36.04M & 29.3 & - & - \\
		SSD-ResNet101 \cite{dssd} & - & 31.2 & 50.4 & 33.3 \\
		DSSD \cite{dssd} & - & 33.2 & 53.3 & 35.2 \\
		\hline
		MobileNetV2$^\dagger$ \cite{mobilenetv2} & 16.68M & 31.5 & 50.9 & 33.5 \\
		RetinaNet$^\dagger$ \cite{focal} & 37.97M & 35.6 & 55.6 & 38.4 \\
		\hline
		Hit-Detector & 33.05M & 36.9 & 55.2 & 39.5 \\
		\bottomrule[1pt]
	\end{tabular}
\end{table}

\section{Conclusion}
In this work, we propose a hierarchical trinity architecture search scheme to address the problem that incomplete searching of detector would cause the inconsistency between different components and lead to the suboptimal performance. We reveal that different components prefer different operations and thus screen three sub search spaces to improve the searching efficiency. Then we search for all components of object detectors based on the sub search spaces in an end-to-end manner. The searched architecture, namely Hit-Detector, achieves state-of-the-art performance on COCO benchmark without bells and whistles.\\
\textbf{Acknowledgement} This work of J. Guo and C. Zhang is supported by the National Nature Science Foundation of China under Grant 61671027 and the National Key R\&D Program of China under Grant 2017YFB1002400, and C. Xu is supported by the Australian Research Council under Project DE180101438.
\section{Appendix}

In this supplementary material, we list the search space and corresponding sub search spaces for detector trinity in details, and we show the qualitative results of our Hit-Detector compared with other state-of-the-art methods. 

\subsection{Search Space}
The whole search space consists of $N=32$ different operation candidates in our experimental setting. We list the candidates bellow:

\begin{itemize}
\vspace{-0.1cm}
    \item ir\_k3\_d1\_e1 \inlineitem ir\_k3\_d1\_e3 \inlineitem ir\_k3\_d1\_e6
\vspace{-0.2cm}
    \item ir\_k3\_d2\_e1 \inlineitem ir\_k3\_d2\_e3 \inlineitem ir\_k3\_d2\_e6
\vspace{-0.2cm}
    \item ir\_k3\_d3\_e1 \inlineitem ir\_k3\_d3\_e3 \inlineitem ir\_k3\_d3\_e6
\vspace{-0.2cm}
    \item ir\_k5\_d1\_e1 \inlineitem ir\_k5\_d1\_e3 \inlineitem ir\_k5\_d1\_e6
\vspace{-0.2cm}
    \item ir\_k5\_d2\_e1 \inlineitem ir\_k5\_d2\_e3 \inlineitem ir\_k5\_d2\_e6
\vspace{-0.2cm}
    \item ir\_k5\_d3\_e1 \inlineitem ir\_k5\_d3\_e3 \inlineitem ir\_k5\_d3\_e6
\vspace{-0.2cm}
    \item ir\_k7\_d1\_e1 \inlineitem ir\_k7\_d1\_e6
\vspace{-0.2cm}
    \item sep\_k3\_d1  \,\, \inlineitem sep\_k3\_d2    \,\,  \inlineitem sep\_k3\_d3
\vspace{-0.2cm}
    \item sep\_k5\_d1  \,\, \inlineitem sep\_k5\_d2    \,\,  \inlineitem sep\_k5\_d3
\vspace{-0.2cm}
    \item conv\_k3\_d1 \inlineitem conv\_k3\_d2 \inlineitem conv\_k3\_d3
\vspace{-0.2cm}
    \item conv\_k5\_d1 \inlineitem conv\_k5\_d2 \inlineitem conv\_k5\_d3
\end{itemize}

where ``ir", ``sep" and ``conv" indicates the  inverted residual block, separable block and convolution block, respectively, ``k" indicates the kernel size, ``d" indicates the dilation rate, ``e" indicates the expansion rate of inverted residual block.  

\subsection{Sub search space}

In our experiments, we set $N_b=N_n=N_h=8$, and we list the top-8 candidate operations for \textbf{backbone}:
\begin{itemize}
\vspace{-0.1cm}
    \item ir\_k3\_d1\_e3 \inlineitem ir\_k3\_d1\_e6 \inlineitem ir\_k3\_d2\_e3
\vspace{-0.2cm}
    \item ir\_k5\_d1\_e3 \inlineitem ir\_k5\_d1\_e3 \inlineitem ir\_k5\_d2\_e6
\vspace{-0.2cm}
    \item ir\_k5\_d3\_e6 \inlineitem ir\_k7\_d1\_e6
\end{itemize}

The top-8 candidate operations for \textbf{neck}:
\begin{itemize}
\vspace{-0.1cm}
    \item conv\_k3\_d3 \inlineitem conv\_k5\_d1 \inlineitem ir\_k3\_d2\_e1
\vspace{-0.2cm}
    \item ir\_k5\_d1\_e3 \inlineitem sep\_k3\_d1 \,\, \inlineitem sep\_k3\_d3
\vspace{-0.2cm}
    \item sep\_k5\_d2  \,\, \inlineitem sep\_k5\_d3
\end{itemize}

The top-8 candidate operations for \textbf{head}:
\begin{itemize}
\vspace{-0.1cm}
    \item ir\_k3\_d1\_e3 \inlineitem ir\_k3\_d1\_e6 \inlineitem ir\_k3\_d2\_e6
\vspace{-0.2cm}
    \item ir\_k5\_d1\_e3 \inlineitem ir\_k5\_d1\_e6 \inlineitem ir\_k7\_d1\_e6
\vspace{-0.2cm}
    \item conv\_k3\_d1 \inlineitem conv\_k5\_d1
\end{itemize}

\subsection{Qualitative results}
We show the qualitative results of our Hit-Detector. We randomly sample some images from the COCO minival and show detection results with confidence bigger than 0.5. First column is the results of FPN \cite{fpn}, second column is the results of NAS-FPN \cite{nasfpn} implemented by ourselves, third column is the results of DetNAS \cite{detnas}, and the fourth column is the results of our Hit-Detector. Different box colors indicate different object categories.

\begin{table*}
     \hspace{-0.25cm}
	\setlength{\tabcolsep}{1mm}
     \begin{tabular}{cccc}
     FPN \cite{fpn} & NAS-FPN \cite{nasfpn} & DetNAS \cite{detnas} & Hit-Detector \\
     \includegraphics[width=0.245\linewidth]{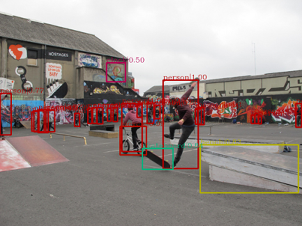}
      &\includegraphics[width=0.245\linewidth]{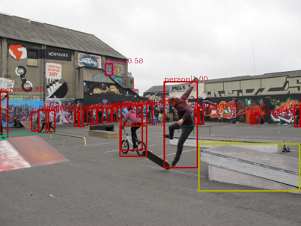}
      &\includegraphics[width=0.245\linewidth]{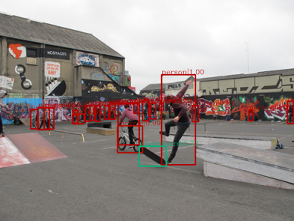}
      &\includegraphics[width=0.245\linewidth]{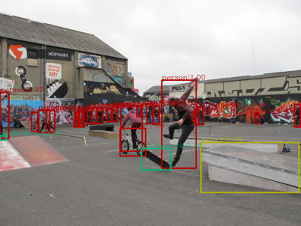}\\
     \includegraphics[width=0.245\linewidth]{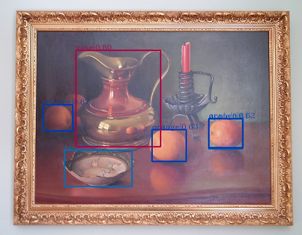}
      &\includegraphics[width=0.245\linewidth]{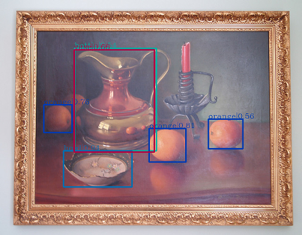}
      &\includegraphics[width=0.245\linewidth]{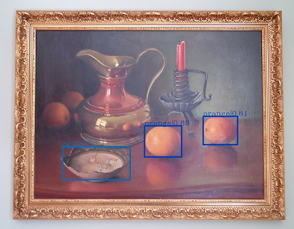}
      &\includegraphics[width=0.245\linewidth]{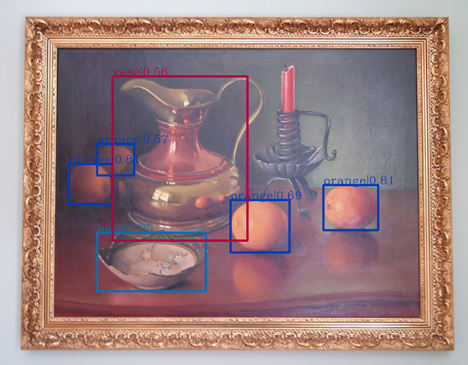}\\
     \includegraphics[width=0.245\linewidth]{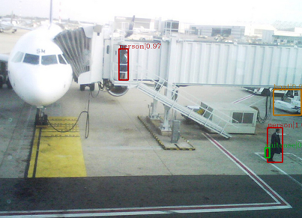}
      &\includegraphics[width=0.245\linewidth]{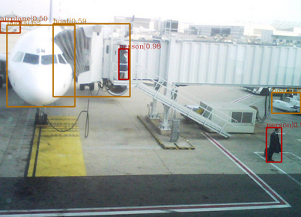}
      &\includegraphics[width=0.245\linewidth]{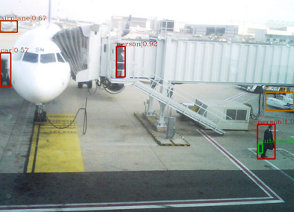}
      &\includegraphics[width=0.245\linewidth]{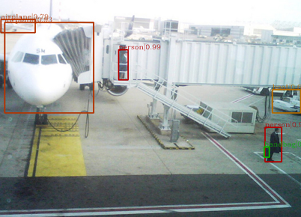}\\
     \includegraphics[width=0.245\linewidth]{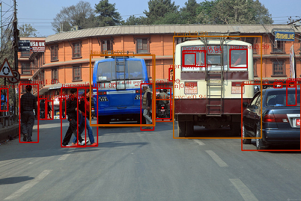}
      &\includegraphics[width=0.245\linewidth]{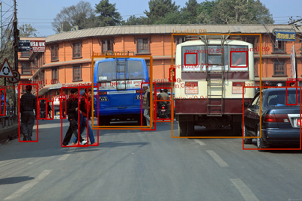}
      &\includegraphics[width=0.245\linewidth]{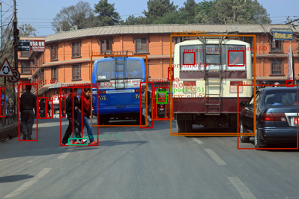}
      &\includegraphics[width=0.245\linewidth]{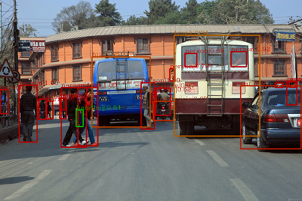}\\
     \includegraphics[width=0.245\linewidth]{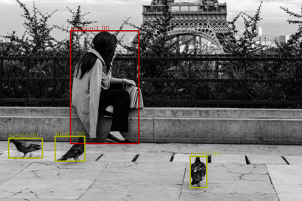}
      &\includegraphics[width=0.245\linewidth]{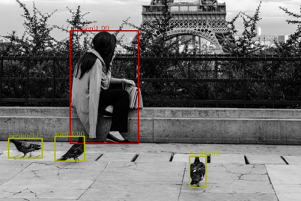}
      &\includegraphics[width=0.245\linewidth]{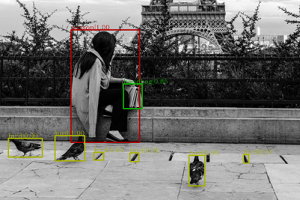}
      &\includegraphics[width=0.245\linewidth]{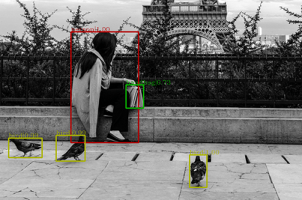}\\
     \includegraphics[width=0.245\linewidth]{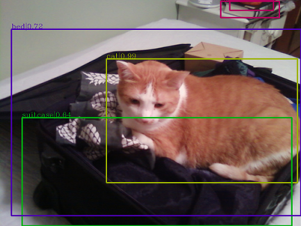}
      &\includegraphics[width=0.245\linewidth]{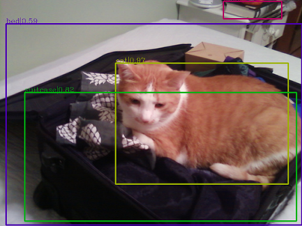}
      &\includegraphics[width=0.245\linewidth]{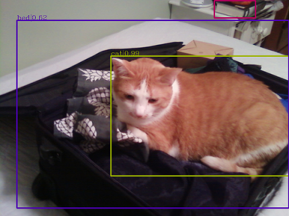}
      &\includegraphics[width=0.245\linewidth]{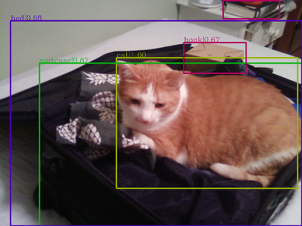}\\
     \includegraphics[width=0.245\linewidth]{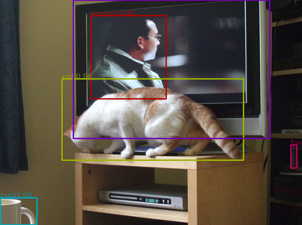}
      &\includegraphics[width=0.245\linewidth]{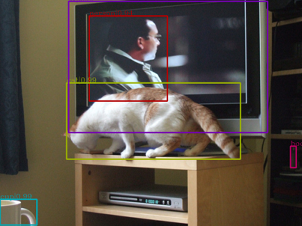}
      &\includegraphics[width=0.245\linewidth]{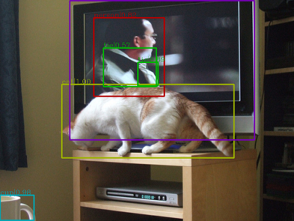}
      &\includegraphics[width=0.245\linewidth]{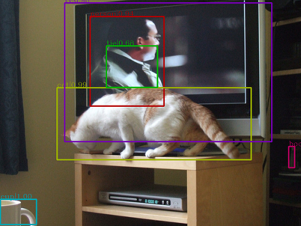}\\
      \end{tabular}
      \label{table:demo}
\end{table*}

\begin{table*}
     \hspace{-0.25cm}
	\setlength{\tabcolsep}{1mm}
     \begin{tabular}{cccc}
     FPN \cite{fpn} & NAS-FPN \cite{nasfpn} & DetNAS \cite{detnas} & Hit-Detector \\
     \includegraphics[width=0.245\linewidth]{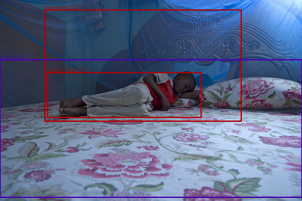}
      &\includegraphics[width=0.245\linewidth]{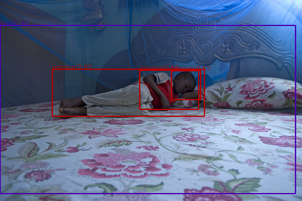}
      &\includegraphics[width=0.245\linewidth]{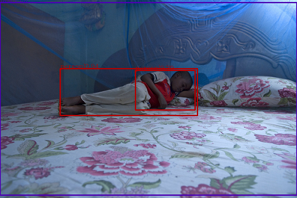}
      &\includegraphics[width=0.245\linewidth]{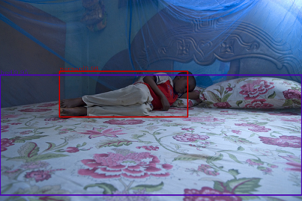}\\
     \includegraphics[width=0.245\linewidth]{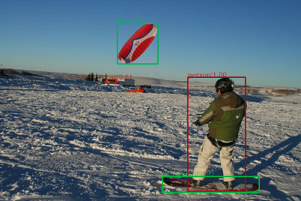}
      &\includegraphics[width=0.245\linewidth]{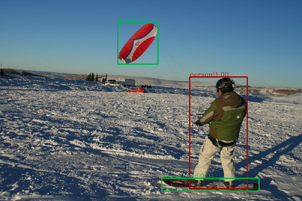}
      &\includegraphics[width=0.245\linewidth]{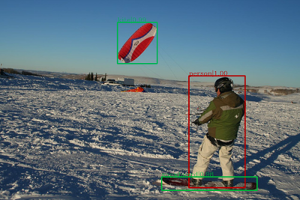}
      &\includegraphics[width=0.245\linewidth]{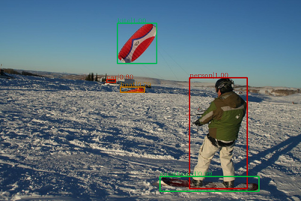}\\
     \includegraphics[width=0.245\linewidth]{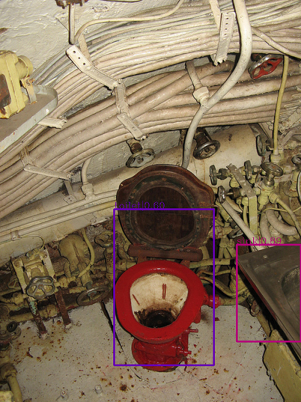}
      &\includegraphics[width=0.245\linewidth]{supp_fig/nasfpn_138.png}
      &\includegraphics[width=0.245\linewidth]{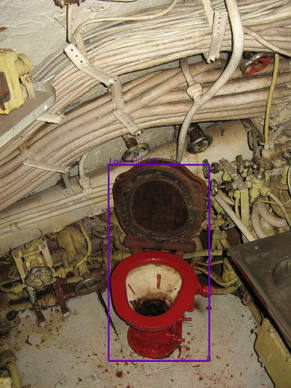}
      &\includegraphics[width=0.245\linewidth]{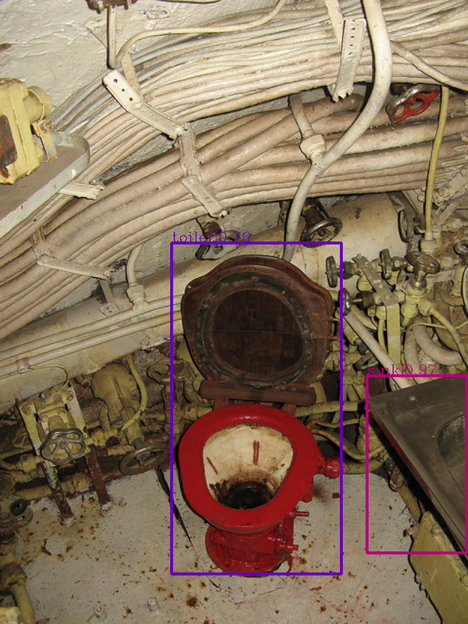}\\
     \includegraphics[width=0.245\linewidth]{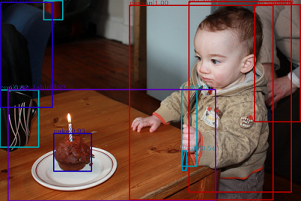}
      &\includegraphics[width=0.245\linewidth]{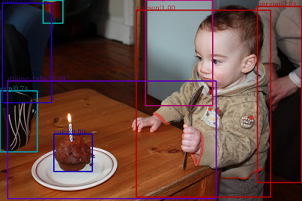}
      &\includegraphics[width=0.245\linewidth]{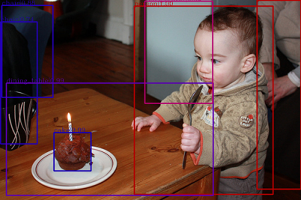}
      &\includegraphics[width=0.245\linewidth]{supp_fig/hit_139.png}\\
     \includegraphics[width=0.245\linewidth]{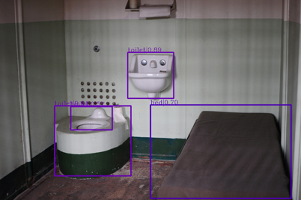}
      &\includegraphics[width=0.245\linewidth]{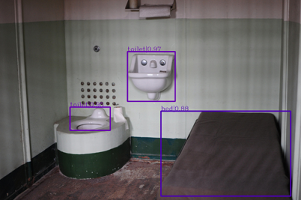}
      &\includegraphics[width=0.245\linewidth]{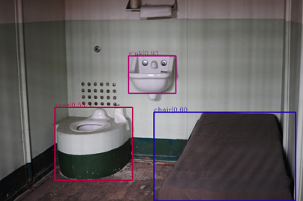}
      &\includegraphics[width=0.245\linewidth]{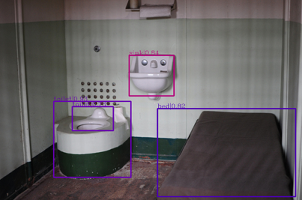}\\
     \includegraphics[width=0.245\linewidth]{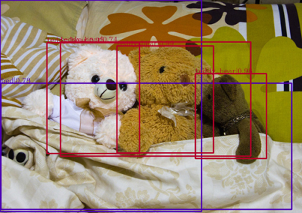}
      &\includegraphics[width=0.245\linewidth]{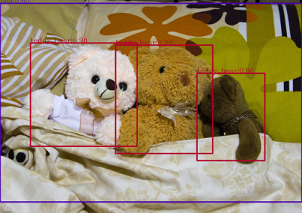}
      &\includegraphics[width=0.245\linewidth]{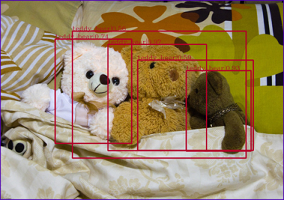}
      &\includegraphics[width=0.245\linewidth]{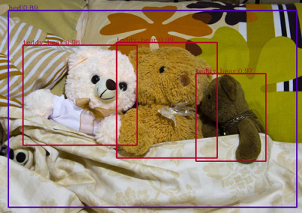}\\
      \end{tabular}
      \label{table:demo}
\end{table*}

{\small
	\bibliographystyle{ieee_fullname}
	\bibliography{egbib}
}

\end{document}